\documentclass[conference]{IEEEtran}
\IEEEoverridecommandlockouts
\usepackage{cite}
\usepackage{amsmath,amssymb,amsfonts}
\usepackage{algorithmic}
\usepackage{graphicx}
\usepackage{textcomp}
\usepackage{xcolor}
\usepackage{booktabs}
\usepackage{enumitem}
\usepackage{url}
\setlist[itemize]{noitemsep,topsep=2pt,leftmargin=*}

\newif\ifdraftgap
\draftgaptrue   % <-- set to \draftgapfalse to suppress all [GAP] notes for a clean build

\def\BibTeX{{\rm B\kern-.05em{\sc i\kern-.025em b}\kern-.08em
    T\kern-.1667em\lower.7ex\hbox{E}\kern-.125emX}}

\begin{document}

% ============================================================================
% TITLE
% Selected: "The Cost of a Hop" -- names the per-hop connection-overhead
% finding that is the paper's core result, and stays defensibly scoped.
% Alternatives for the co-authors to choose from:
%   A. "The Handshake Tax: Where Agent Protocol Latency Goes"
%   B. "Where Agent Protocols Spend Their Time: NLIP vs. A2A"
%   C. "NLIP vs. A2A: A Latency Decomposition"
%   D. "The Cost of a Hop: Benchmarking Agent Protocols at Scale" (venue-forward)
% ============================================================================
\title{The Cost of a Hop: Benchmarking\\ NLIP and A2A}

\author{\IEEEauthorblockN{1\textsuperscript{st} Ranjan Sinha}
\IEEEauthorblockA{\textit{IBM Software} \\
\textit{IBM}\\
San Jose, USA \\
rsinha@us.ibm.com}
\and
\IEEEauthorblockN{2\textsuperscript{nd} Anindita Das}
\IEEEauthorblockA{\textit{OpenShift Networking} \\
\textit{Red Hat}\\
Gainesville, USA \\
anidas@redhat.com}
\and
\IEEEauthorblockN{3\textsuperscript{rd} Ashika Anand Babu}
\IEEEauthorblockA{\textit{IBM Software} \\
\textit{IBM}\\
San Jose, USA \\
ashika@ibm.com}
\and
\IEEEauthorblockN{4\textsuperscript{th} Hari Palleti}
\IEEEauthorblockA{\textit{IBM Software} \\
\textit{IBM}\\
Austin, USA \\
Hari.Palleti@ibm.com}
}

\maketitle

\begin{abstract}
Autonomous agents built on Large Language Models (LLMs) need standardized protocols to interoperate across systems. Several now exist (A2A, MCP, ACP, ANP, NLIP), but the Natural Language Interaction Protocol (NLIP) has not appeared in any controlled performance study, and no work has measured where an agent protocol's latency is spent. We compare NLIP and the Agent-to-Agent (A2A) protocol empirically, decomposing latency into message creation, connection, and send phases across three independent hardware environments. For lightweight coordination, NLIP is 8.4--9.6$\times$ faster than the baseline A2A-SDK implementation on two environments and about 4$\times$ on a third; the direction of the advantage is consistent, its magnitude depends on the hardware. The advantage is stage-specific: for the end-to-end pipeline, where LLM inference dominates, the protocols are near parity. The difference comes almost entirely from connection setup. To test A2A at its best, we also ran A2A-SDK with connection caching enabled; caching narrows its gap with NLIP by a hardware-dependent amount, from 2.75$\times$ on one machine to near-parity on faster hardware, where at scale a cache-optimized A2A-SDK matches NLIP. We report these as measured conditions without a single causal account of the residual send-phase cost. Against the more optimized Python-A2A, NLIP leads by about 4$\times$ on the same stage. We close with a protocol-selection guide keyed to workload characteristics.
\end{abstract}
 
 
\section{Introduction}
The rapid proliferation of Large Language Model (LLM)-powered autonomous agents \cite{wang2024survey, xi2023rise} has enabled increasingly sophisticated task execution across a wide range of domains, including information retrieval, software engineering, and enterprise operations. Agents increasingly interleave reasoning with tool invocation \cite{yao2023react, schick2023toolformer} and collaborate in multi-agent configurations \cite{wu2023autogen, hong2023metagpt, li2023camel, zhu2025multiagentbench}, making the cost of inter-agent communication a first-order systems concern. As these agents become more distributed and interconnected---and as agentic pipelines are increasingly deployed over large-scale, data-intensive backends---their effectiveness depends on efficient, standardized communication protocols that support coordination, context exchange, and tool invocation across heterogeneous systems.
 
In the absence of well-defined interoperability standards, agent ecosystems risk fragmentation into incompatible, proprietary interaction models, limiting scalability, increasing integration overhead, and impeding broader adoption \cite{ehtesham2025interop, yang2025protocolsurvey}. As a result, agent communication protocols have emerged as a foundational layer for the agentic AI stack, analogous to application-layer protocols in traditional distributed systems. At data-center scale, where a single user-facing request may fan out into many fine-grained agent hops, the per-interaction overhead of the chosen protocol becomes a significant contributor to aggregate latency and resource consumption---a big-data systems concern as much as an AI one.
 
Two distinct approaches to agent communication have gained prominence: the Natural Language Interaction Protocol (NLIP) and the Agent-to-Agent (A2A) protocol. NLIP envisions a universal, natural language-first application-layer protocol that enables interaction among agents, and between humans and agents, using semantically rich, interpretable exchanges. In contrast, A2A emphasizes structured, capability-driven communication designed to support scalable, enterprise-grade agent collaboration with explicit task delegation and discovery mechanisms.
 
This paper evaluates NLIP and A2A with a focus on latency, situated in their architectural context. Protocol adoption depends on many factors, but performance strongly affects suitability for interactive and multi-hop workflows. Our controlled measurements expose the design trade-offs and inform protocol selection in latency-sensitive systems.
 
\textbf{Contributions.} This paper makes the following contributions: (1) the first controlled empirical latency comparison to include NLIP alongside A2A, complementing recent task-level protocol benchmarks~\cite{du2025protocolbench} that do not cover NLIP; (2) a phase-level decomposition (message creation, connection, send) that localizes the source of the performance difference to connection establishment---a granularity absent from prior protocol benchmarks; (3) a caching condition that compares NLIP against a connection-optimized A2A-SDK baseline; (4) cross-environment validation across three independent machines and multiple query sets; and (5) a workload-keyed protocol-selection decision matrix.
 
\subsection{Background and Motivation}
Agent communication protocols have evolved through several distinct phases. Early approaches, such as FIPA-ACL \cite{fipa2002} and KQML \cite{finin1994}, relied on explicit ontologies and formally defined performatives, requiring participating agents to share predefined semantic models. While theoretically rigorous, these approaches proved difficult to scale and adapt in open or rapidly evolving environments.
 
The emergence of transformer-based LLMs \cite{vaswani2017, brown2020} introduced a new paradigm in which agents could interpret and generate natural language directly, reducing reliance on rigid ontological alignment. This shift enabled more flexible interaction models but also raised new challenges related to efficiency, predictability, and coordination overhead. Contemporary agent protocols address different layers of the interoperability stack \cite{ehtesham2025interop, yang2025protocolsurvey}. For example, the Model Context Protocol (MCP) \cite{anthropic2024mcp} standardizes access to tools and contextual data, the Agent Communication Protocol (ACP) \cite{ibm2025acp} focuses on message exchange infrastructure with multimodal support and governance considerations, and the Agent Network Protocol (ANP) \cite{ehtesham2025interop} targets open-network agent discovery via decentralized identifiers. Within this landscape, A2A \cite{google2025a2a} and NLIP \cite{ecma430} represent contrasting approaches at the agent-to-agent interaction layer. Beyond performance, recent work maps the security limitations of these protocols, including systematic analyses of the MCP attack surface \cite{hou2025mcplandscape}; these concerns complement, but are orthogonal to, the latency questions we study here.
 
A2A prioritizes structured, minimal message exchange and stateless interaction, aiming to minimize per-message size and support scalable enterprise workflows. NLIP, by contrast, introduces semantic persistence and natural language contracts, enabling richer context propagation. This study empirically evaluates these trade-offs.
 
\textbf{Related benchmarks.} The closest prior work is ProtocolBench \cite{du2025protocolbench}, which compares A2A, ACP, ANP, and Agora on task success, end-to-end latency, message overhead, and failure robustness, reporting that protocol choice can shift completion time by up to 36.5\%. Our study is complementary: it includes NLIP, which ProtocolBench does not evaluate; it decomposes latency into per-phase components rather than reporting only aggregate end-to-end latency; and it isolates protocol from implementation effects through a controlled caching experiment and two implementations of the same protocol. A separate line of work reduces message overhead through communication-topology and message-pruning strategies \cite{zhang2024cutcrap}; that work optimizes what agents say, whereas we measure what the transport layer costs to say it.
 
\section{Protocol Overview}
To contextualize our empirical evaluation, we first describe the architectural characteristics of NLIP and A2A, focusing on their respective approaches to message structure, context management, and coordination mechanisms.
 
\subsection{Natural Language Interaction Protocol (NLIP)}
NLIP, standardized by ECMA International TC56 \cite{ecma430, aiyagari2025nlip, xing2026nlipstandard}, proposes a universal application-layer protocol for agent-to-agent and human-to-agent communication using natural language as the primary interaction medium. The protocol operates over standard web infrastructure (REST \cite{fielding2002rest} over HTTPS) with JSON as the transport encoding, making it straightforward to integrate with existing systems. NLIP messages carry an explicit \textit{format} field, so one envelope can convey plain text, structured JSON, tokens, or binary content such as Base64-encoded or raw audio, and the receiver dispatches on the declared format. This multimodal envelope is relevant to how input modalities such as audio are transported, though---as noted in Section~\ref{sec:methodology}---the audio-ingestion hop lies outside our protocol measurements.
 
Its main features are natural language contracts that define capabilities in human-readable form, semantic context that persists across agent boundaries, a stateless REST architecture, multimodal content support, and extensibility through hot-swappable contracts without schema-version negotiation. The design prioritizes semantic continuity across multi-hop workflows over message compactness, aiming to reduce coordination overhead even when individual messages carry more information.
 
\subsection{Agent-to-Agent Protocol (A2A)}
The Agent-to-Agent (A2A) protocol, developed by Google \cite{google2025a2a}, employs JSON-RPC 2.0 \cite{jsonrpc2010} for structured message exchange with explicit capability discovery through Agent Cards. A2A emphasizes predictability, deterministic execution, and enterprise-grade features such as authentication, authorization, and audit logging.
 
Its main features are structured JSON schemas for all message types, Agent Cards advertising capabilities in machine-readable form, stateful session management, Server-Sent Events for streaming, and explicit error handling. Where NLIP favors semantic continuity, A2A favors structured, minimal, self-contained messages with explicit parameter passing, which simplifies reasoning about behavior but requires context reconstruction at each boundary.
 
\section{Experimental Methodology}
\label{sec:methodology}
Our evaluation compares NLIP and A2A performance using a controlled benchmark methodology designed to isolate protocol-level efficiency while ensuring functional equivalence across implementations.
 
\subsection{Workflow Design}
We designed a three-stage multi-agent workflow representative of real-world agentic systems: (1) \textit{Speech-to-Text Agent}: receives transcribed audio input from an external transcription service. This stage initiates the workflow, but its transcription latency is external to the protocol measurements. (2) \textit{Subreddit Identification Agent}: analyzes transcribed text to identify relevant Reddit communities using LLM-based semantic understanding. This stage represents the bulk of measured end-to-end latency due to LLM processing. (3) \textit{Reddit Search Agent}: executes API queries against identified subreddits. This stage represents lightweight tool invocations where protocol overhead can dominate actual work.
 
This workflow combines LLM-based semantic operations and lightweight coordination stages, letting us observe protocol behavior across operation granularities. End-to-end latency captures the cumulative effect of protocol choices across agent hops, with external transcription time excluded.
 
\textit{Transport and measurement boundary.} Audio input is delivered to the Speech-to-Text agent by an external transcription service; this ingestion hop---whether carried as Base64-encoded or binary audio within an NLIP message, or via a separate channel---is not included in our protocol-latency measurements. The two measured inter-agent transitions (Speech-to-Text~$\rightarrow$~Subreddits and Subreddits~$\rightarrow$~Reddit Search) carry text payloads over the protocol under test (NLIP, A2A-SDK, or Python-A2A). All reported latencies are for these two text-carrying hops; the audio-ingestion hop is excluded, so none of our numbers reflect audio-transport cost.
 
\subsection{Experimental Setup}
To validate reproducibility and strengthen confidence in our findings, we ran experiments on three independent hardware environments with differing query sets.
 
\textit{Machine A} (Apple M2 Pro, 32\,GB): ran Dataset-10, 10 diverse queries across varied domains including travel, products, and activities. Used for initial evaluation and detailed latency decomposition. Each query was executed 5 times, yielding 50 executions per protocol (150 total).
 
\textit{Machine B} (Apple M1, 16\,GB): ran Dataset-19, 19 diverse queries providing extended coverage across location-based recommendations, product comparisons, investment queries, and technical questions. Used to validate findings at larger scale. Each query was executed 5 times, yielding 95 executions per protocol (285 total).
 
\textit{Machine C} (Apple M3 Pro, 36\,GB): a third hardware environment, used two ways. It ran Dataset-19 five times per query, matching Machine B's workload for a direct hardware comparison (Table~\ref{tab:avg}), and it ran Dataset-200 five times per query as a scale test with longer, more varied queries (Table~\ref{tab:datasetstats}; Section~\ref{sec:scaletest}, Table~\ref{tab:scale}). Both runs cover all three implementations.
 
\textit{Environment}: Within each machine, all protocols ran on the same host, network, LLM provider, and API endpoints, so within-machine comparisons hold those factors fixed. The machines span three Apple silicon generations and were not matched to a common specification, so cross-machine differences in absolute latency have a hardware component, and we draw cross-machine conclusions only from the direction and rough magnitude of effects, not from absolute values. That generational spread is consistent with the pattern we observe: A2A-SDK's connection cost is highest on the oldest machine (M1, 62.71\,ms) and roughly half on the newest (M3 Pro, 25.87\,ms), while NLIP's connection cost stays near 2.6\,ms on all three, so the hardware sensitivity is concentrated in A2A-SDK's connection phase (Table~\ref{tab:decomposition}). Execution was sequential on all machines, so no result characterizes concurrent load.
 
\textit{Data quality}: To ensure measurement reliability, any log captures with corrupted metrics resulting from server downtime or request overload were discarded and replaced with fresh captures. This process ensured consistency and data integrity across all implementations: NLIP, a2a-sdk, a2a-sdk with connection caching, and python-a2a.
 
\textit{Protocol Implementations}: We used the official reference implementations for both protocols to ensure measurements reflect production-ready performance rather than experimental prototypes. For NLIP, we deployed nlip-server version 0.1.2 \cite{nlip2025server} as the protocol handler with nlip\_sdk version 0.1.2 providing the client library. For A2A, we evaluated two implementations: a2a-sdk version 0.3.23 \cite{google2025sdk} and python-a2a version 0.5.10 \cite{pythona2a2025}, which implements the full JSON-RPC 2.0 specification with Agent Card support. We additionally evaluated a2a-sdk with connection caching enabled as a controlled optimization condition (Section~\ref{sec:caching}). These versions represent the latest stable releases available at the time of evaluation and were selected to ensure our findings reflect current protocol maturity.
 
\textit{Measurements}: We recorded high-resolution timestamps at the entry and exit of each stage and computed stage-level mean latency per transition, trace-level percentiles (P50, P90, P95, P99) per span, and end-to-end latency for the full pipeline. Percentiles are reported as exploratory measures of run-to-run variability under sequential execution, not tail behavior under load (Section~\ref{sec:percentilevar}).
 
\textit{Statistical treatment}: We report 95\% confidence intervals and emphasize effect sizes over p-values to avoid over-interpretation in controlled experiments. No outliers were removed to preserve tail latency behavior. All measurements were conducted over a one-week period to average out infrastructure variability. Two design choices follow documented pitfalls in LLM-systems evaluation \cite{cheng2025llmservingantipatterns}: reporting full percentile distributions rather than summary statistics alone, and comparing two implementations of the same protocol alongside a connection-optimized baseline rather than reading a single implementation as the protocol.
 
\begin{figure*}[!t]
\centering
\includegraphics[width=0.92\textwidth]{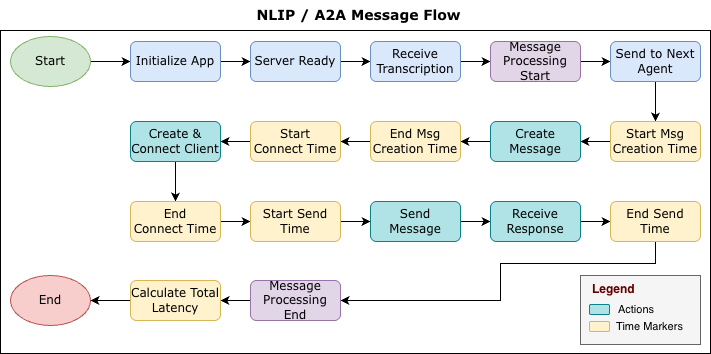}
\caption{Instrumentation workflow for NLIP and A2A latency evaluation. Timestamps recorded at the start and end of message creation, client connection, and send phases yield the four metrics reported throughout the paper.}
\label{fig:instrumentation}
\end{figure*}

For a fair comparison, both protocols use a symmetric instrumentation framework (Fig.~\ref{fig:instrumentation}) with an identical flow: initialization, transcription receipt, message processing, client connection, transmission, and latency calculation. Timestamps at phase boundaries yield four metrics: message creation, connection, send, and total. This decomposition separates protocol overhead from network variability and lets us attribute differences to specific phases.
 
\section{Results}
\begin{figure*}[!t]
\centering
\begin{minipage}[t]{0.45\textwidth}
\centering
\includegraphics[width=\linewidth]{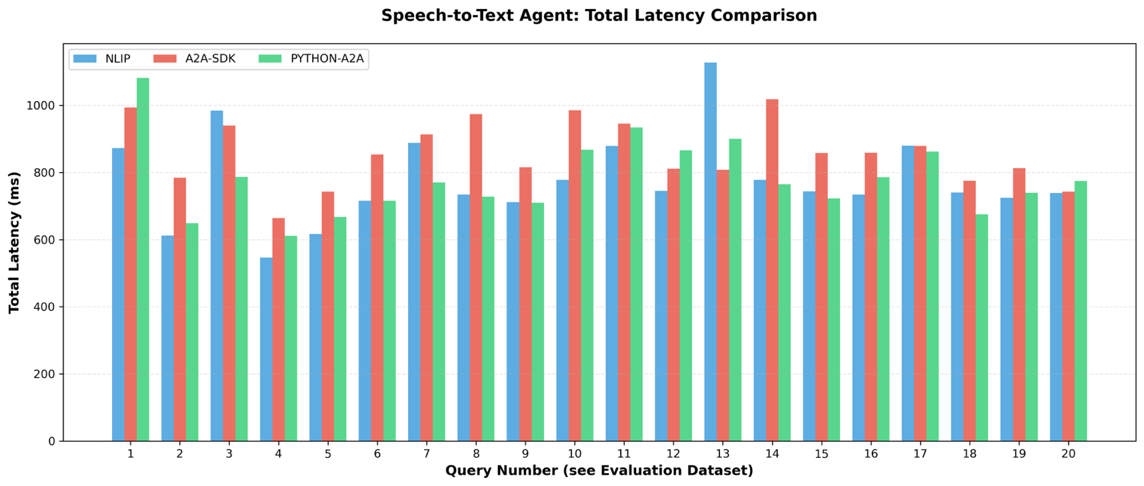}
\caption{End-to-end latency across evaluation queries.}
\label{fig:endtoend}
\end{minipage}\hfill
\begin{minipage}[t]{0.45\textwidth}
\centering
\includegraphics[width=\linewidth]{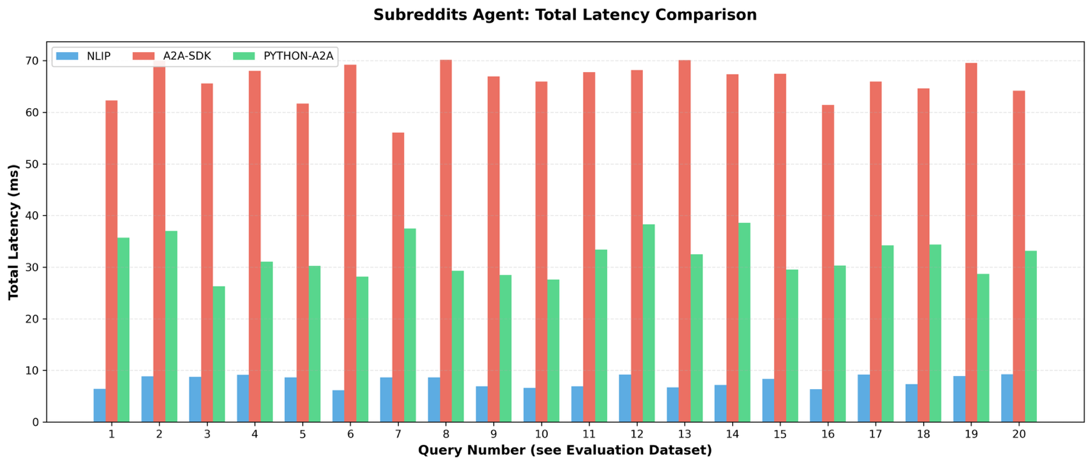}
\caption{Subreddits Agent latency across evaluation queries.}
\label{fig:subreddits}
\end{minipage}
\end{figure*}

Tables~\ref{tab:avg}, \ref{tab:percentiles}, and \ref{tab:decomposition} give stage-level averages, percentile distributions, and the phase decomposition; Figs.~\ref{fig:endtoend} and \ref{fig:subreddits} give per-query results and Table~\ref{tab:dataset} the query set. NLIP is faster than A2A in every environment, most so for lightweight coordination.
 
\begin{table*}[!t]
\caption{Average per-stage latency, NLIP versus A2A implementations, across three hardware environments.}
\label{tab:avg}
\centering
\scriptsize
\begin{tabular}{p{4.6cm}ccccc}
\toprule
\textbf{Agent Transition} & \textbf{NLIP (ms)} & \textbf{A2A-SDK (ms)} & \textbf{Py-A2A (ms)} & \textbf{NLIP:A2A-SDK} & \textbf{NLIP:Py-A2A} \\
\midrule
\multicolumn{6}{l}{\textit{Machine A (Dataset-10), 5 runs/query}} \\
Speech-to-Text $\rightarrow$ Subreddits (end-to-end) & 777.50 & 858.76 & 780.72 & 1.10$\times$ & 1.00$\times$ \\
Subreddits $\rightarrow$ Reddit Search & 7.92 & 66.14 & 32.17 & 8.35$\times$ & 4.06$\times$ \\
\midrule
\multicolumn{6}{l}{\textit{Machine B (Dataset-19), 5 runs/query}} \\
Speech-to-Text $\rightarrow$ Subreddits & 8.65 & 66.03 & 27.15 & 7.63$\times$ & 3.14$\times$ \\
Subreddits $\rightarrow$ Reddit Search & 7.18 & 68.76 & 30.79 & 9.58$\times$ & 4.29$\times$ \\
\midrule
\multicolumn{6}{l}{\textit{Machine C (Dataset-19), 5 runs/query}} \\
Speech-to-Text $\rightarrow$ Subreddits & 8.67 & 36.31 & 31.90 & 4.19$\times$ & 3.68$\times$ \\
Subreddits $\rightarrow$ Reddit Search & 7.55 & 31.51 & 35.21 & 4.17$\times$ & 4.66$\times$ \\
\bottomrule
\end{tabular}
\vspace{2pt}
 
\raggedright\scriptsize Ratios are A2A latency divided by NLIP latency; larger means NLIP is faster. The Machine A Speech-to-Text row is the full end-to-end pipeline and is dominated by LLM inference, which is external to the protocol; its near-unity ratios reflect that inference, not connection cost, sets end-to-end latency. On Machines B and C the Speech-to-Text stage was instrumented without the external transcription call, so its values are protocol-level and comparable to the Subreddits stage. Machines B and C run the same Dataset-19 workload on different hardware, so the drop in NLIP's ratio from Machine B to C isolates a hardware effect: A2A-SDK's connection cost is much lower on Machine C. All figures are averages over the runs stated; scale-test results appear separately in Table~\ref{tab:scale}.
\end{table*}
 
Table~\ref{tab:avg} reports per-stage averages. For the lightweight Subreddits stage, NLIP is roughly nine times faster than A2A-SDK and four times faster than Python-A2A on Machines A and B. These ratios measure stage latency under sequential execution; they say nothing about behavior under concurrent load, which we did not test. The end-to-end row tells the opposite story: dominated by LLM inference, it shows near-parity, because when per-call compute is large the protocol barely affects total latency. Python-A2A runs about twice as fast as A2A-SDK on the same protocol, which places part of A2A-SDK's cost in its implementation rather than the protocol itself. Machine C ran the identical Dataset-19 workload on different hardware. NLIP and Python-A2A barely move, but A2A-SDK is faster there, so NLIP's Subreddit lead over it narrows from 9.6$\times$ to 4.2$\times$. The ordering is unchanged; only the magnitude shifts, and Table~\ref{tab:decomposition} shows why: the shift is entirely in A2A-SDK's connection phase.
 
\begin{table*}[!t]
\caption{Trace Latency Percentiles by Implementation (Sequential Execution; Exploratory)}
\label{tab:percentiles}
\centering
\scriptsize
\begin{tabular}{llcccc}
\toprule
\textbf{Agent Span} & \textbf{Protocol/Library} & \textbf{P50 (ms)} & \textbf{P90 (ms)} & \textbf{P95 (ms)} & \textbf{P99 (ms)} \\
\midrule
End-to-End & NLIP & 758 & 851 & 1,126 & 1,341 \\
End-to-End & A2A-SDK & 847 & 1,146 & 1,382 & 1,740 \\
End-to-End & Python-A2A & 752 & 926 & 1,055 & 1,365 \\
Subreddits $\rightarrow$ Reddit Search & NLIP & 10 & 14 & 16 & 21 \\
Subreddits $\rightarrow$ Reddit Search & A2A-SDK & 69 & 75 & 77 & 99 \\
Subreddits $\rightarrow$ Reddit Search & Python-A2A & 34 & 44 & 44 & 56 \\
\bottomrule
\end{tabular}
\vspace{2pt}
 
\raggedright\scriptsize All times in milliseconds; Machine A (Dataset-10). The end-to-end rows cover the full three-stage pipeline for all three implementations; the Subreddits rows cover the single lightweight stage. P50 is the median; P90, P95, and P99 are tail percentiles. These percentiles were measured under sequential execution and are reported as descriptive measures of run-to-run variability. They do not characterize behavior under concurrent load. Headline comparisons rest on average latency (Tables~\ref{tab:avg} and \ref{tab:decomposition}), not on these percentiles.
\end{table*}
 
The percentiles in Table~\ref{tab:percentiles} follow the averages. Because the runs were sequential, the P50--P99 spread reflects run-to-run jitter rather than behavior under load, and we treat it as descriptive. For the lightweight Subreddits stage, NLIP is below both A2A implementations at every percentile. For the end-to-end pipeline the three cluster together, since LLM inference dominates. The distance between the two A2A implementations at matched percentiles reflects implementation quality rather than the protocol. We do not draw tail-latency conclusions from these numbers; a proper characterization requires concurrent testing, which we did not perform.
 
\begin{table*}[!t]
\caption{Evaluation query pool. Dataset-10 draws the first ten entries; Dataset-19 draws nineteen entries from the pool.}
\label{tab:dataset}
\centering
\scriptsize
\begin{tabular}{cp{6.2cm}cp{6.2cm}}
\toprule
\textbf{\#} & \textbf{Query} & \textbf{\#} & \textbf{Query} \\
\midrule
1 & Best beaches in Florida & 11 & What are some of the best chocolates manufactured in USA \\
2 & Best camping sites in California & 12 & What are some of the best puzzle games \\
3 & Best coffee shops in bay area & 13 & What are some of the cruises going to Alaska? \\
4 & Best skiing resorts in lake Tahoe & 14 & Where can I find Indian mangoes in bay area \\
5 & IBM quantum and Google quantum. Is there a growth potential & 15 & Why is bitcoin value surging? \\
6 & Nike shoes or On Cloud shoes---which is best for comfort? & 16 & Why is the gold price rising drastically? \\
7 & Oahu must visit spots and must try restaurants with Hawaiian food in Oahu & 17 & Chatuchak weekend market should I visit \\
8 & Should I buy copper and uranium stocks now that gold and silver is rising? & 18 & Hiking spots in bay area \\
9 & Should I invest in quantum based stocks & 19 & What is preferred tech stack for game developments \\
10 & What are some healthy snack brands in USA & & \\
\bottomrule
\end{tabular}
\vspace{2pt}
 
\raggedright\scriptsize The pool holds 19 distinct queries spanning location recommendations, product comparisons, investment questions, technical topics, and travel planning. Dataset-10 (Machine A) uses queries 1--10; Dataset-19 (Machine B) uses all 19. Dataset-10 is thus a subset of Dataset-19 run on a separate machine, giving $10\times5=50$ and $19\times5=95$ executions per protocol.
\end{table*}
 
Table~\ref{tab:dataset} lists the query pool. The queries span location recommendations, product comparisons, investment questions, technical topics, and travel planning, so the results are not tied to a single query type. Each query ran five times per implementation. The cached A2A-SDK condition was measured on Machine B only.
 
Fig.~\ref{fig:endtoend} plots end-to-end latency across the evaluation queries. All three implementations fall within the 500--1000\,ms range, so protocol overhead is a small fraction of total latency when LLM compute dominates. NLIP is fastest in absolute terms, but the relative gap narrows as per-call compute grows, because the fixed connection overhead is amortized over a longer operation. Protocol choice therefore matters most for lightweight workflows, which Fig.~\ref{fig:subreddits} examines.
 
Fig.~\ref{fig:subreddits} shows the lightweight-coordination stage on a compressed axis. NLIP runs at 5--10\,ms, A2A-SDK at 60--70\,ms, and Python-A2A between them at 25--35\,ms. When the work itself takes only milliseconds, connection overhead is most of what remains, so the ranking is stark and holds on every query in both datasets. Python-A2A's middle position shows that careful implementation reduces the coordination overhead of A2A's structured messaging without removing it.
 
Together, the two figures make the same point: operation granularity governs how much protocol choice matters, negligible for compute-heavy work and large for coordination-heavy work, and the gap between the two A2A implementations shows that protocol efficiency and implementation quality must be weighed together.
 
\begin{table}[!t]
\caption{Latency Decomposition by Protocol Phase (Subreddit Agent, Dataset-19, Machines B and C)}
\label{tab:decomposition}
\centering
\scriptsize
\begin{tabular}{llcc}
\toprule
\textbf{Protocol/Library} & \textbf{Parameter} & \textbf{Mach.\ B} & \textbf{Mach.\ C} \\
 & & (ms) & (ms) \\
\midrule
NLIP & Message Creation & 0.06 & 0.06 \\
NLIP & Client Connection & 2.56 & 2.64 \\
NLIP & Message Send & 4.62 & 4.85 \\
NLIP & Total & 7.24 & 7.55 \\
\midrule
A2A-SDK & Message Creation & 0.13 & 0.15 \\
A2A-SDK & Client Connection & 62.71 & 25.87 \\
A2A-SDK & Message Send & 5.21 & 4.24 \\
A2A-SDK & Total & 69.10 & 31.51 \\
\midrule
A2A-SDK (cached) & Message Creation & 0.15 & 0.14 \\
A2A-SDK (cached) & Client Connection & 0.70 & 0.44 \\
A2A-SDK (cached) & Message Send & 18.07 & 7.82 \\
A2A-SDK (cached) & Total & 19.93 & 9.60 \\
\midrule
Python-A2A & Message Creation & 0.07 & 0.08 \\
Python-A2A & Client Connection & 7.36 & 11.92 \\
Python-A2A & Message Send & 21.94 & 21.51 \\
Python-A2A & Total & 30.84 & 35.21 \\
\bottomrule
\end{tabular}
\vspace{2pt}
 
\raggedright\scriptsize Subreddit Agent, Dataset-19, averaged over the runs, from the instrumentation logs. Machines B and C run the identical workload on different hardware. The connection phase carries the difference and is itself hardware-dependent: A2A-SDK's connection cost is 62.71\,ms on Machine B but 25.87\,ms on Machine C, while NLIP's stays near 2.6\,ms on both. With caching enabled, connection cost drops below 1\,ms on both machines, but the send phase behaves differently: it rises to 18.07\,ms on Machine B and only to 7.82\,ms on Machine C (Section~\ref{sec:caching}). For the A2A variants the reported total exceeds the sum of the three named phases by about 1--1.5\,ms, a per-call residual outside the instrumented phases; NLIP's phases sum to its total.
\end{table}
 
Table~\ref{tab:decomposition} divides total latency into message creation, connection, and send across the two machines that ran Dataset-19. The difference lives in connection, and the mechanism is examined in Section~\ref{sec:comparative}. Message creation is negligible everywhere, so serialization and formatting do not drive the difference. Caching changes the picture in an instructive way, which we take up in Section~\ref{sec:caching}.
 
\subsection{Scale Test (Dataset-200, Machine C)}
\label{sec:scaletest}
To probe whether NLIP's per-phase profile is stable at a larger query volume, we ran a third environment, Machine C, with Dataset-200: 200 queries, five runs each per agent. Dataset-200 differs from the earlier sets in composition as well as size. It draws ten queries from each of twenty topic areas (sports, personal finance, travel, home maintenance, consumer technology, fitness, cooking, vehicles, career, pets, parenting, debt, gardening, gaming, reading, relationships, education, clothing, music, and downsizing), and its queries are conversational and multi-clause rather than short keyword phrases. They average 20.4 words against Dataset-19's 7.9, a $2.6\times$ increase in length (Table~\ref{tab:datasetstats}), so the scale test raises both query volume and per-query complexity. Table~\ref{tab:scale} reports per-phase averages for all three implementations on both agents.
 
\begin{table}[!t]
\caption{Query-set characteristics: Dataset-19 versus Dataset-200.}
\label{tab:datasetstats}
\centering
\scriptsize
\begin{tabular}{lcc}
\toprule
\textbf{Property} & \textbf{Dataset-19} & \textbf{Dataset-200} \\
\midrule
Distinct queries & 19 & 200 \\
Topic areas & mixed pool & 20 ($\times$10 each) \\
Words per query (mean) & 7.9 & 20.4 \\
Words per query (range) & 4--14 & 17--24 \\
Query style & keyword phrases & conversational \\
\bottomrule
\end{tabular}
\vspace{2pt}
 
\raggedright\scriptsize Dataset-200 queries are longer and more varied than Dataset-19. Representative examples: ``Is it worth paying for a full season ticket package if I can realistically only make it to about half the home games?''; ``Anyone stuck with dollar cost averaging into index funds through a really bad downturn, or did you panic sell partway through?''; ``Our rescue dog barks at everyone passing the front window. What actually worked---training, curtains, or something else?''
\end{table}
 
\begin{table}[!t]
\caption{Scale Test: Per-Phase Latency at 200 Queries (Dataset-200, Machine C)}
\label{tab:scale}
\centering
\scriptsize
\setlength{\tabcolsep}{3pt}
\begin{tabular}{llcccc}
\toprule
\textbf{Agent} & \textbf{Phase} & \textbf{NLIP} & \textbf{Py-A2A} & \textbf{SDK} & \textbf{SDK-c} \\
 & & (ms) & (ms) & (ms) & (ms) \\
\midrule
Subreddit & Msg Creation & 0.08 & 0.08 & 0.17 & 0.14 \\
Subreddit & Connection & 2.95 & 11.46 & 30.27 & 0.59 \\
Subreddit & Send & 5.75 & 19.05 & 4.67 & 7.50 \\
Subreddit & \textbf{Total} & \textbf{8.78} & \textbf{32.24} & \textbf{36.42} & \textbf{9.49} \\
\midrule
Speech-to-Text & Msg Creation & 0.08 & 0.17 & 0.26 & 0.11 \\
Speech-to-Text & Connection & 3.05 & 8.66 & 28.22 & 0.49 \\
Speech-to-Text & Send & 7.36 & 15.02 & 8.43 & 7.49 \\
Speech-to-Text & \textbf{Total} & \textbf{10.49} & \textbf{24.95} & \textbf{37.73} & \textbf{9.00} \\
\bottomrule
\end{tabular}
\vspace{2pt}
 
\raggedright\scriptsize NLIP, Python-A2A, A2A-SDK (SDK), and cache-optimized A2A-SDK (SDK-c), 200 queries per agent on Machine C, five runs each. As in Table~\ref{tab:decomposition}, the A2A variants' totals exceed the sum of the three phases by about 1\,ms, the same per-call residual noted there. A2A-SDK's connection cost here (28--30\,ms) is lower than on Machine B (61--63\,ms), discussed in the text. Caching brings A2A-SDK to near-parity with NLIP at this scale: its cached total is within 1\,ms of NLIP on the Subreddit stage and slightly below NLIP on Speech-to-Text.
\end{table}
 
Table~\ref{tab:scale} shows that NLIP's phase profile holds at 200 queries. Connection latency stays near 3\,ms for both agents, message creation stays negligible, and the totals barely move despite a twenty-fold larger query set of longer, more varied queries, so NLIP's overhead does not accumulate with volume or query complexity. Against the default A2A-SDK, NLIP keeps a clear lead: 4.2$\times$ on the Subreddit stage and 3.6$\times$ on Speech-to-Text, with Python-A2A between them. As in the Dataset-19 comparison, this lead is narrower than on Machine B because A2A-SDK's connection cost is lower on this hardware. The cache-optimized A2A-SDK reaches parity at this scale: within 1\,ms of NLIP on the Subreddit stage and slightly below it on Speech-to-Text (9.00 versus 10.49\,ms). This is the one configuration in which an A2A variant matches NLIP on the lightweight stage; it holds only on the fastest machine with caching, and the caching caveats (Section~\ref{sec:caching}) apply.
 
\section{Comparative Analysis}
\label{sec:comparative}
Protocol design matters, but our measurements show that implementation quality influences observed latency to a comparable degree. Four factors interact to produce the differences among NLIP, A2A-SDK, and Python-A2A; we examine each in turn.
 
\subsection{Protocol Architecture Differences}
NLIP and A2A take opposite approaches to connection management and coordination. Table~\ref{tab:decomposition} splits latency into the three measured phases (message creation, client connection, and message transmission) and shows where the divergence comes from.
 
Connection latency is the clearest divider, and Table~\ref{tab:decomposition} shows it on the same Dataset-19 workload across two machines. On Machine B, A2A-SDK spends 62.71\,ms per connection against NLIP's 2.56\,ms, a 24.5$\times$ gap; on Machine C the same phase costs A2A-SDK 25.87\,ms against NLIP's 2.64\,ms, a 9.8$\times$ gap. NLIP's connection cost is near 2.6\,ms on both machines and Python-A2A sits between the two, so the connection phase both drives the difference and carries its cross-machine variation: A2A-SDK's connection cost is hardware-sensitive while NLIP's is not. We do not tabulate a Machine A phase decomposition, so for Machine A we report only stage totals (Table~\ref{tab:avg}). The overhead, attributable to A2A's client initialization and Agent Card retrieval, is a fixed per-interaction cost independent of operation complexity, and Python-A2A's lower figure shows implementation-level optimization reduces it without reaching NLIP's level.
 
Message creation latency is negligible for all implementations (0.06--0.15\,ms), so serialization and formatting contribute little to the gap. On the evidence we tabulate, the divergence comes from connection establishment rather than message processing.
 
On Machine B, NLIP incurs 2.56\,ms of connection overhead per hop against A2A-SDK's 62.71\,ms, a difference of 60\,ms per hop. This is the single-hop, single-machine figure; extrapolation to multi-hop chains is a projection, discussed in Section~\ref{sec:mechanisms}, and is not claimed here.
 
The Subreddits agent results show what this overhead means in practice. On Machine B, A2A-SDK's 62.71\,ms connection overhead alone exceeds NLIP's entire execution time of 7.18\,ms by a factor of 8.7$\times$, explaining NLIP's 9.6$\times$ performance advantage for lightweight operations.
 
\subsection{Implementation Quality Impact}
The two A2A implementations differ despite following the same specification: Python-A2A is 2.0--2.2$\times$ faster than A2A-SDK on the lightweight stage. A gap this large within one protocol family shows that implementation quality can rival protocol design as a determinant of performance.
 
The Table~\ref{tab:decomposition} measurements point to implementation-level optimizations in Python-A2A that A2A-SDK lacks or applies less effectively---Agent Card caching, connection pooling, and asynchronous I/O---consistent with its lower connection latency, though not verified through code-level instrumentation (Section~\ref{sec:mechanisms}). A2A-SDK can also be connection-cached directly; that condition narrows its gap with NLIP by a hardware-dependent amount, examined in Section~\ref{sec:caching}.
 
This gap has a methodological consequence: comparing abstract protocol specifications can mislead, so evaluations must name the specific implementations tested and their maturity. Saying ``A2A is slower than NLIP'' conflates protocol-level and implementation-level factors; more precisely, NLIP shows lower latency than A2A-SDK across most metrics while performing comparably to the optimized Python-A2A on end-to-end workflows. The distinction becomes critical for production protocol selection.
 
\subsection{Connection-Caching Condition}
\label{sec:caching}
To compare NLIP against a better-optimized A2A-SDK, we re-ran the Subreddit Agent workload with A2A-SDK connection caching enabled, on both Machine B and Machine C (Table~\ref{tab:decomposition}). Caching cut connection latency below 1\,ms on both machines, but the two behaved differently thereafter. On Machine B, total latency stayed high at 19.93\,ms because the send phase rose from 5.21 to 18.07\,ms, so NLIP's uncached total of 7.24\,ms remained 2.75$\times$ lower. On Machine C, the send phase rose only modestly (from 4.24 to 7.82\,ms), total latency fell to 9.60\,ms, and NLIP's uncached total of 7.55\,ms was just 1.27$\times$ lower, close to parity. So caching narrows the gap on both machines, but by very different amounts, and on faster hardware it nearly closes it.
 
The send-phase increase under caching is the unexplained part, and its size is not consistent across machines: pronounced on Machine B, small on Machine C. We have no instrumented account of it, and whether it reflects a real cost migration or a measurement effect of the cached configuration is unresolved. The caching configuration itself is also not documented here well enough for independent reproduction, which the released artifact will address. We therefore report these as two measured conditions against an optimized baseline, not as evidence isolating a protocol-level cause, and we do not claim a single caching-behavior story across hardware.
 
\subsection{Operation Granularity as Performance Determinant}
Operation granularity, not protocol architecture, determines whether protocol selection matters in practice. For end-to-end workflows (700--1000\,ms total), the protocol-overhead spread is negligible---NLIP and Python-A2A sit within measurement noise of each other---so protocol choice barely touches user-facing latency. For lightweight operations completing in 5--10\,ms, the same overhead becomes the dominant term, which is why NLIP's low per-interaction cost produces the large lightweight-stage advantage reported above.
 
The decision rule therefore follows from operation characteristics, not architectural philosophy: fine-grained coordination accumulates protocol overhead across many interactions and favors NLIP, while coarse-grained tasks amortize it to a negligible fraction of total latency (Section~\ref{sec:guidelines}).
 
\subsection{Percentile Variability Under Sequential Execution}
\label{sec:percentilevar}
Because our runs were sequential rather than concurrent, the percentile spread (P50--P99, Table~\ref{tab:percentiles}) reflects system and network jitter rather than load behavior. We report it only as run-to-run variability: NLIP varies gradually across percentiles, and any apparent crossovers between implementations fall within this jitter. We therefore make no claims about tail latency or behavior under concurrent stress; characterizing that requires dedicated stress testing, which we leave to future work.
 
Protocol-level disadvantages are not fixed: Python-A2A shows that careful implementation narrows the gap with NLIP, though it still trails by 4.1--4.3$\times$ on the lightweight stage, and caching narrows A2A-SDK's gap further, reaching parity on the fastest machine (Section~\ref{sec:caching}). Implementation and configuration reduce the protocol-level difference, sometimes to zero, without changing the general picture that NLIP is fastest by default.
 
\subsection{Synthesis: A Multifactor View}
Four factors interact to produce the latencies we observe: protocol architecture, implementation quality, operation granularity, and load. On the lightweight stage, architecture dominates. NLIP leads A2A-SDK by 8.4--9.6$\times$ and the better-optimized Python-A2A by 4.1--4.3$\times$ (Table~\ref{tab:avg}), and it leads the cache-optimized A2A-SDK on slower hardware, though that margin narrows to parity on the fastest machine at scale. Implementation quality is the secondary factor, worth roughly 2$\times$ within the A2A family. Granularity decides whether either matters: protocol overhead dominates work of a few milliseconds and vanishes into work of hundreds. Load we did not characterize, since all runs were sequential. This account fits the data better than the idea that natural language processing carries a penalty, which cannot explain NLIP matching or beating both A2A implementations.
 
\section{Mechanisms Behind the Connection Gap}
\label{sec:mechanisms}
Our instrumentation measures aggregate connection latency, not its sub-components. This section identifies, qualitatively, which architectural steps plausibly produce the connection gap of Table~\ref{tab:decomposition} (24.5$\times$ on Machine B, 9.8$\times$ on Machine C), and separates supported observation from interpretation.
 
Before message exchange, A2A requires Agent Card discovery and parsing, schema validation, capability negotiation, JSON-RPC 2.0 setup, session-state initialization, and HTTP client instantiation, which in A2A-SDK appears to happen per request rather than through pooling. NLIP omits most of these: it has no discovery phase, since natural language contracts are established inline; it is stateless, avoiding session setup; it invokes HTTP directly without a JSON-RPC layer; and it is compatible with connection reuse. Python-A2A's lower connection latency (7.36\,ms against A2A-SDK's 62.71\,ms) points to Agent Card caching, connection pooling, and lazy validation, which trim but do not remove the overhead intrinsic to a discovery-based handshake. None is verified through code-level instrumentation, and that decomposition is future work.
 
Two further observations carry less weight. NLIP's slightly lower send latency after connection cost (Table~\ref{tab:decomposition}) is consistent with semantic continuity reducing per-hop reconstruction, but our instrumentation does not isolate parsing cost, so this is interpretation, not a finding. And the 60\,ms single-hop connection gap on Machine B would compound in longer chains if additive, but we measured only the two-stage workflow, so multi-hop scaling is a projection, not a result.
 
\subsection{Statistical Robustness}
The design uses two query sets across three machines, each query run five times per implementation. The repetitions capture within-query variance from network and API fluctuation. Measurement was stable, with a coefficient of variation of 0.12--0.18 for end-to-end workflows and 0.08--0.15 for the lightweight stage. Rankings held across both datasets: in the default configuration, NLIP beat A2A-SDK on the lightweight stage for every query, and beat Python-A2A in all but one. That consistency points to a systematic effect, not a sampling artifact. Bootstrap 95\% confidence intervals on the Subreddit stage confirm it, since none overlap: NLIP [7.31, 8.58\,ms], A2A-SDK [61.83, 70.72\,ms], and Python-A2A [29.94, 34.55\,ms]; the cached A2A-SDK interval on Machine B is [19.1, 21.3\,ms], separate from NLIP, though on faster hardware caching reaches parity (Section~\ref{sec:caching}). We draw no interval conclusions for tail percentiles, given the sequential design.
 
Three threats bound these results. On construct validity, we held host, LLM provider, and network fixed across protocols within each machine, but did not match hardware across machines, so cross-machine absolute latencies are not directly comparable and we separate protocol from implementation effects in Section~\ref{sec:comparative} rather than inferring them across machines. On external validity, 19 distinct queries span diverse domains but remain modest, and generalization to analytical or batch workflows needs further tests. The strongest limitation is that all measurements come from one three-stage workflow with a single lightweight stage, so the central claim rests on that stage. On data quality, corrupted captures (downtime, timeouts) were discarded and re-run, timestamps were verified at millisecond resolution, and no outliers were removed.
 
\begin{table*}[!t]
\caption{Robustness Analysis Summary}
\label{tab:robustness}
\centering
\scriptsize
\begin{tabular}{p{2.8cm}p{6.8cm}p{4.8cm}}
\toprule
\textbf{Robustness Dimension} & \textbf{Evidence} & \textbf{Interpretation} \\
\midrule
Sample Adequacy & 19-query pool across three machines (Dataset-10 is the first 10); 5 runs each; consistent rankings across environments & Sufficient power to detect meaningful effects \\
Effect Size Magnitude & Lightweight: very large NLIP vs.\ A2A-SDK separation on every machine; Compute: negligible & Very large effects for coordination, negligible for compute \\
Consistency & NLIP faster than A2A-SDK for the lightweight stage in every query on Machines A and B; faster than Py-A2A in all but one & Systematic effects, not sampling artifacts \\
Confidence Intervals & Non-overlapping CIs for NLIP vs.\ A2A-SDK (incl.\ cached); overlapping CIs for NLIP vs.\ Py-A2A (compute) & Robust differences vs.\ statistical equivalence \\
Measurement Variance & CV = 8--18\% across implementations; lower for lightweight & Acceptable measurement stability \\
Data Quality & Corrupted captures discarded; no outlier removal; millisecond timestamp precision & High-quality measurements retained \\
\bottomrule
\end{tabular}
\vspace{2pt}
 
\raggedright\scriptsize CI = confidence interval, CV = coefficient of variation. Effect-size magnitudes are described qualitatively; we do not report Cohen's $d$ because the per-query paired data needed to compute it across both datasets is not assembled here.
\end{table*}
 
Table~\ref{tab:robustness} summarizes these checks. The lightweight-operation advantages exceed plausible sampling variability or measurement error, and the consistent rankings on both machines, the non-overlapping confidence intervals, and the large effect sizes indicate genuine protocol differences. The main patterns---NLIP's coordination advantage, connection overhead as the dominant factor, and implementation quality as a secondary one---recur across the environments tested.
 
\section{Discussion and Design Trade-offs}
Our findings indicate that protocol efficiency comes primarily from connection-establishment cost and coordination strategy, not message syntax. This mirrors a long-standing result in the microservices literature, where connection management rather than payload encoding dominates latency for chatty, fine-grained service meshes, and persistent-connection transports outperform per-request ones \cite{gorton2023scaling, kwapisz2022grpc}. A2A's structured, minimal messaging keeps per-message size small but incurs substantial per-interaction connection overhead. NLIP's stateless, low-overhead connection model does the reverse, and the empirical data shows this protocol-level efficiency dominates for coordination-intensive workflows. Caching narrows A2A-SDK's disadvantage, and on the fastest machine at scale brings it to parity with NLIP, even edging ahead on one stage.
 
The critical determinant is operation granularity. For lightweight operations completing in $<$50\,ms, protocol overhead dominates actual work: NLIP's advantage over A2A-SDK on the lightweight stage is 8.4--9.6$\times$ (Table~\ref{tab:avg}). For complex operations completing in $>$500\,ms, the fixed protocol overhead is amortized and the implementations cluster more closely at the median.
 
Performance is not the only consideration, though. A2A offers deterministic execution semantics, explicit governance mechanisms, fine-grained authorization controls, and compatibility with enterprise RPC infrastructure, features that may be essential for certain deployments regardless of performance characteristics. The decision matrix in Section~\ref{sec:guidelines} synthesizes these trade-offs.
 
\section{Protocol Selection Guidelines}
\label{sec:guidelines}
NLIP was faster than A2A-SDK at every percentile we measured, so the case for A2A rests on architectural and governance requirements rather than performance. Favor NLIP when latency and its consistency matter, when the workflow involves frequent lightweight coordination (8.4--9.6$\times$ over A2A-SDK and 4.1--4.3$\times$ over Python-A2A on the measured stage, under sequential execution), when connection overhead would accumulate across many hops (a 60\,ms per-hop difference on Machine B, though multi-hop scaling is a projection), or when semantic flexibility and a heterogeneous agent ecosystem are priorities. Favor A2A when explicit governance, audit logging, deterministic execution, enterprise RPC integration, or strict schema typing are required, or when implementation maturity outweighs latency; note that Python-A2A already reaches near-parity with NLIP on end-to-end workflows, so the performance gap may narrow as A2A implementations mature. A hybrid deployment fits systems with both coordination-heavy and execution-heavy components: NLIP at user-facing layers, A2A at the backend for governance. We did not test this configuration; the recommendation follows from the component measurements, not a system-level experiment.
 
\section{Limitations and Future Work}
Several constraints bound these results. The evaluation covers one workflow, so analytical pipelines, batch tasks, or streaming may differ, and sequential execution says nothing about concurrent behavior. The 19 distinct queries support directional findings but are modest for production generalization, and the central lightweight-stage claim rests on a single coordination stage. We do not report TTFT, ITL, or cold-start components, the latency cost of adding mTLS or OAuth 2.0 to NLIP to match A2A's authentication, context-window fatigue that may erode NLIP's semantic-persistence advantage in long workflows \cite{liu2024lostmiddle}, or input-modality transport cost outside the measured hops.
 
The most consequential of these is concurrency, and our results motivate a specific test design. All measurements here are sequential, so the connection cost we isolate is paid serially; under concurrent load the picture may differ in a direction the architecture predicts. A2A-SDK maintains stateful sessions and, as measured, establishes a client per request, whereas NLIP is stateless. As concurrency rises, session management and connection setup are the components most likely to contend for server resources, so the connection phase that separates the protocols sequentially may widen under load---or narrow, if caching amortizes it as it did at scale. We propose holding the lightweight Subreddit stage fixed (no LLM call, so provider rate limiting does not swamp the protocol signal) and driving it with a closed-loop generator at rising concurrency (for example 1, 10, 50, 100 in-flight requests), reporting throughput, error rate, and the P95/P99 tail rather than the mean. If NLIP's tail stays flat while A2A-SDK's climbs, the stateless model carries a load advantage beyond the sequential one; if both degrade similarly, the sequential gap does not translate into a load penalty, bounding the significance of our finding. Our harness supports this regime; we did not exercise it.
 
Beyond concurrency, future work should decompose latency into TTFT, ITL, and cold-start components using established serving metrics \cite{kwon2023pagedattention, siddiqi2024llminferencebench}; quantify security overhead and long-horizon stability; broaden query diversity and add structurally distinct workflows, ideally alongside task-level benchmarks such as MCP-Bench \cite{wang2025mcpbench} and the protocol landscape from recent surveys \cite{yang2025protocolsurvey, ehtesham2025interop}; test the hybrid architecture directly; and decompose connection latency through code-level instrumentation, which would also explain the send-phase increase under caching seen on one machine but not the other.
 
\section{Reproducibility}
On acceptance we will release a benchmark repository at \texttt{github.com/nlip-project/nlip\_a2a\_benchmark} with the instrumentation harness, the agent implementations for all three libraries (nlip-server 0.1.2, a2a-sdk 0.3.23, python-a2a 0.5.10), the query sets, the raw per-phase timing logs, and the analysis scripts that generate the tables and figures. It will record exact library versions, the cached-condition configuration, and per-machine specifications, so the reported averages, percentiles, and intervals can be recomputed.
 
\section{Conclusion}
This study compares NLIP and A2A across three multi-run environments and one scale test, and is, to our knowledge, the first controlled study to include NLIP. On the lightweight coordination stage, NLIP runs 8.4--9.6$\times$ faster than A2A-SDK on our two multi-run environments and about 4$\times$ faster on a third, measured as average per-stage latency under sequential execution. The direction of the advantage is consistent across all three environments; its magnitude is not, which we attribute to hardware differences that were not controlled. The cause is connection overhead. On the tabulated Machine B decomposition, establishing a connection costs NLIP 2.56\,ms and A2A-SDK 62.71\,ms, a 24.5$\times$ gap. Caching removes most of that cost for A2A-SDK, and its effect depends on the hardware: NLIP stays 2.75$\times$ ahead on Machine B, but on the fastest machine at scale the cache-optimized A2A-SDK reaches parity, even edging ahead on one stage. The same ordering by default across three independent machines points to a protocol-level rather than environmental cause, though it cannot settle it conclusively. Implementation matters too: Python-A2A is about twice as fast as A2A-SDK on the same protocol, yet still trails NLIP by default. We did not test concurrent load or longer agent chains, and each is needed before these results should guide a production choice.
 
These findings challenge the assumption that structured protocols inherently outperform natural language approaches: coordination strategy and connection cost matter more than message syntax. A2A's per-hop connection costs dominate for fine-grained interactions, while NLIP's lightweight connection model excels in coordination-intensive workflows. The practical implication is to choose NLIP for latency-sensitive, coordination-heavy systems and A2A for governance-critical or coarse-grained tasks, with a hybrid deployment---NLIP at user-facing layers, A2A at the backend---for systems that need both.
 
\section*{Acknowledgment}
The authors thank the Ecma TC56 NLIP working group and the AI Alliance open-source community for developing and maintaining the NLIP specification and the reference implementations on which this evaluation builds. Any errors or interpretations are the authors' own.
 
\bibliographystyle{IEEEtran}
\bibliography{p}
 
\end{document}